\documentclass{article}

\PassOptionsToPackage{sort&compress}{natbib}
\usepackage[preprint]{corl_2026}

\usepackage[utf8]{inputenc}
\usepackage[T1]{fontenc}
\usepackage{url}
\usepackage{booktabs}
\usepackage{amsfonts}
\usepackage{amsmath,amssymb}
\usepackage{amsthm}
\usepackage{nicefrac}
\usepackage{microtype}
\usepackage{graphicx}
\usepackage{multirow}
\usepackage{wrapfig}
\usepackage{float}
\usepackage{algorithm}
\usepackage{algpseudocode}
\usepackage{enumitem}
\usepackage{xcolor}

\newcommand{\rd}[1]{\textcolor{red}{#1}}

\hypersetup{
   pdfsubject={}, 
   pdftitle={Perceptive Behavior Foundation Model: Adapting Human Motion Priors to Robot-Centric Terrain},
   pdfauthor={Zifan Wang, Yizhao Li, Teli Ma, Qiang Zhang, Yudong Fan, Hao Xu, Shuo Yang, Junwei Liang},
   pdfkeywords={perceptive humanoid control; behavior foundation models; motion tracking}
}

\title{Perceptive Behavior Foundation Model:\\Adapting Human Motion Priors to Robot-Centric Terrain}

\author{
  Zifan Wang$^{1,2}$ \quad\; Yizhao Li$^{2}$ \quad\; Teli Ma$^{1,2}$ \quad\; Qiang Zhang$^{4}$ \quad\; Yudong Fan$^{2}$\\[3pt]
  \bf Hao Xu$^{5}$ \quad\; Shuo Yang$^{2,\ast}$ \quad\; Junwei Liang$^{1,3,\ast}$
  \AND
  \normalfont
  $^{1}$The Hong Kong University of Science and Technology (Guangzhou) \quad $^{2}$Mondo Robotics\\
  $^{3}$The Hong Kong University of Science and Technology\\
  $^{4}$Artificial General Intelligence Institute, University of Science and Technology of China\\
  $^{5}$School of Artificial Intelligence and Science, Nanjing University\\
  $^{\ast}$Corresponding authors\\
  Project page: \url{https://acodedog.github.io/perceptive-bfm/}
}

\begin{document}
\maketitle

\begin{figure}[H]
\centering
\vspace{-20pt}
\includegraphics[width=\linewidth]{figures/first_page.pdf}
\vspace{-14pt}
\caption{\small \textbf{Single-policy terrain grounding on hardware.} One Perceptive BFM checkpoint tracks diverse flat-ground human-motion commands while adapting footholds, swing clearance, support height, and local contact timing to robot-side obstacles. Locomotion panels use randomly placed terrain; the two phase-critical backflip panels start within the alignment envelope of Sec.~\ref{sec:generality}.}
\label{fig:hero}
\end{figure}

\begin{abstract}
Humanoid behavior foundation models learn reusable whole-body policies from human motion priors, but typically assume that reference motions are compatible with the robot's surroundings. When human and robot environments differ, the reference specifies behavior but not terrain-dependent footholds, clearance, support height, or contact timing. We introduce \emph{Perceptive Behavior Foundation Model} (Perceptive BFM), a motion-conditioned policy pretrained on $16{,}813$ clips ($43.3$\,h) and reused without fine-tuning across unseen references, sources, and terrains. A robot-centric elevation grid supplies terrain information while the raw kinematic reference remains the command interface. Offline \emph{terrain-conformal reference synthesis} (TCRS) adapts foot contacts and support height while preserving upper-body joint references. A blind teacher tracks these synthesized references and transfers its behavior through target-frame action alignment to a raw-reference student, an identity-gated Transformer tracker with zero-initialized terrain gates. In simulation, the deployed checkpoint completes $100\%$ of flat-ground and $94.0\%$ of $1{,}650$ terrain trials across five behaviors and eleven slope, box, and stair cells, versus $19.5$--$23.4\%$ for blind, no-TCRS, and no-map baselines. Map-corruption probes show that the policy conditions on terrain information. On hardware, the same checkpoint executes $35$ motion classes over placed obstacles, including acrobatics and mocap teleoperation with operator--environment mismatch.

\end{abstract}


\section{Introduction}
\label{sec:intro}
Humanoid control is shifting from isolated, task-specific skills toward generalist motion-tracking policies and behavior foundation models: whole-body trackers reproduce diverse motions with a single learned controller~\citep{heOmniH2OUniversalDexterous2024,chengExpressiveWholeBodyControl2024,heHOVERVersatileNeural2024}, and foundation-control systems scale behavior priors, command encoders, policy capacity, and downstream interfaces~\citep{bjorckGR00TN12025,zengBehaviorFoundationModel2025,liBFMZero2025,luoSONIC2025}. Human motion thereby becomes a reusable command interface: one broad prior over human-like whole-body intent instead of a controller per maneuver.

\textbf{What we mean by ``behavior foundation model.''} The term is used in two senses. Meta Motivo~\citep{tirinzoniMetaMotivo2025} and BFM-Zero~\citep{liBFMZero2025} use it for unsupervised reinforcement-learning models that can be prompted zero-shot with arbitrary rewards or tasks. A recent survey~\citep{yuanBFMSurvey2025} also applies it to large motion trackers~\citep{luoSONIC2025,zengBehaviorFoundationModel2025} that earn the label through large-scale pretraining and generalization to unseen motions. We use the term strictly in this second, \emph{motion-conditioned} sense: one policy pretrained on a broad corpus ($16{,}813$ clips, $43.3$\,h from LAFAN1~\citep{harveyLAFAN2020}, AMASS~\citep{mahmoodAMASS2019}, 100STYLE~\citep{mason100STYLE2022}, SNAP, and our own mocap; App.~\ref{app:corpus-dr}) and reused without fine-tuning across unseen references, motion sources, and terrains. We make no reward-prompting or open-ended task claim; Sec.~\ref{sec:generality} reports the source-held-out and scaling evidence behind the label.

This progress exposes a hidden assumption: reference-centric formulations ask how accurately the robot can reproduce the supplied motion, not how that motion should be grounded in the robot's own environment. A flat-ground walking reference does not specify stair footholds: it conveys intent and style, not a terrain-valid trajectory in the robot's world. This is the \emph{operator--environment mismatch}: the human supplies the desired behavior, while the robot must resolve terrain-specific contacts, support height, balance, and timing from its own perception. Perceptive locomotion and parkour policies use elevation maps, depth, and terrain observations to traverse obstacles or sparse footholds~\citep{heAttentionMapEncoding2025,zhuHikingWild2026,zhuangDeepWholeBodyParkour2026,wuPerceptiveHumanoidParkour2026}, but they are organized around traversal skills, motion matching, or system-selected maneuvers rather than preserving an arbitrary human motion command. The missing capability is the perceptual grounding of behavior priors.

We introduce \emph{Perceptive BFM}, a single-policy framework for terrain-aware humanoid motion tracking. The control problem is \emph{perceptive motion tracking}: given a raw kinematic reference, proprioception, and a local elevation grid (no RGB or depth image), generate whole-body actions that are behaviorally faithful and environmentally feasible. The grid supplies only footholds, swing clearance, support height, and local contact timing. Terrain reference adaptation changes foot contacts and support height while preserving upper-body joint references; hand-contact retargeting is outside its scope (Sec.~\ref{sec:limitations}). Training follows a staged \emph{Perceptive Motion Tracking} (PMT) algorithm (Fig.~\ref{fig:teaser}): an offline \emph{terrain-conformal reference synthesis} (TCRS) module converts raw clips and sampled height fields into terrain-consistent supervision while preserving the global phase of the clip; a blind Transformer teacher tracks the synthesized references; a perceptive student receives the original raw reference and the elevation grid and imitates the teacher through target-frame action alignment in its own raw-command frame; and PPO fine-tunes the distilled student. Zero-initialized terrain gates preserve the base tracker at branch insertion; fine-tuning retains the learned gates.

\textbf{Contributions.}
\begin{itemize}[leftmargin=1.2em,itemsep=0pt,topsep=1pt]
    \item \emph{TCRS}, a scalable offline pipeline that converts raw human motion and sampled height fields into terrain-consistent supervision through contact-aware foothold construction, swing optimization, root reconstruction, collision repair, and multi-point foot-contact IK.
    \item \emph{PMT}, a raw-reference teacher--student algorithm that deploys terrain-aware behavior without changing the command interface: a blind teacher tracks TCRS references, and target-frame action alignment transfers its behavior into the raw-reference action frame of a perceptive student.
    \item An identity-gated Transformer for single-policy terrain grounding of broad motion priors, evaluated with pipeline-matched baselines, map-corruption probes, a teacher comparison against a SONIC re-implementation, source-held-out and scaling tests, and $35$ motion classes on hardware (Fig.~\ref{fig:hero}).
\end{itemize}

\begin{figure}[t]
\centering
\includegraphics[width=\linewidth]{figures/pipeline.pdf}
\vspace{-40pt}
\caption{\small \textbf{Perceptive BFM overview.} TCRS synthesizes terrain-conformal references \emph{offline only}. A blind teacher tracks them; the deployed identity-gated student receives the raw reference and a robot-centric elevation grid (``visual terrain input'' in the figure) and learns residual corrections through target-frame action alignment.}
\label{fig:teaser}
\vspace{-10pt}
\end{figure}

\section{Related Work}
\label{sec:related-work}
\paragraph{Generalist humanoid behavior and expressive tracking.}
DeepMimic and AMP established reinforcement learning for tracking motion clips and imitation priors~\citep{pengDeepMimicExample2018,pengAMPAdversarialMotion2021}. Recent humanoid systems scale these ideas to richer corpora and robot embodiments, from teleoperation and universal tracking~\citep{heH2OHumanoidRealtime2024,heOmniH2OUniversalDexterous2024,chengExpressiveWholeBodyControl2024} to broader representations, command encoders, model scale, and residual refinement stacks~\citep{luoPHC2023,luoPULSE2024,heHOVERVersatileNeural2024,fuHumanPlus2024,maRGMT2026,wangOmniXtreme2026,luoSONIC2025}. Behavior foundation models push the same trend toward reusable whole-body policies: Meta Motivo and BFM-Zero learn promptable, reward-free controllers with unsupervised RL~\citep{tirinzoniMetaMotivo2025,liBFMZero2025}, GR00T and \citet{zengBehaviorFoundationModel2025} scale motion-conditioned control~\citep{bjorckGR00TN12025}, and \citet{yuanBFMSurvey2025} survey the space. Perceptive BFM belongs to the motion-conditioned branch and targets a complementary failure mode: broad motion priors encode human intent without specifying terrain-valid contacts in the robot's world.

\paragraph{Perceptive locomotion and terrain-aware whole-body control.}
Terrain-aware legged policies use height maps, depth, or images to traverse obstacles and sparse footholds and to avoid collisions~\citep{agarwalLeggedLocomotion2023,zhuangParkourLegged2023,chengParkourLegged2024,zhuangHumanoidParkour2024,wangBeamDojo2025,sunNowYouSeeThat2026,zhuHikingWild2026,wangOmniPerception2025,wangSafeComfortable2025}, and attention-based map encoders learn to focus on steppable areas~\citep{heAttentionMapEncoding2025}. Whole-body parkour systems integrate exteroception into motion tracking or distill depth policies from expert controllers~\citep{zhuangDeepWholeBodyParkour2026,wuPerceptiveHumanoidParkour2026}. Their task differs from ours: they select, retrieve, or generate a terrain-compatible maneuver, whereas in our setting the user-provided motion is the behavior to preserve and perception must ground that command rather than replace it; this task mismatch is why we report no head-to-head comparison with them (Sec.~\ref{sec:teacher}).

\paragraph{Generated, repaired, and terrain-conditioned references.}
A related interface lets the system choose, generate, or repair a feasible reference: generator--tracker systems synthesize terrain-conditioned motions online~\citep{zhangLearningWholeBodyHumanoid2026}, reference-guided RL modulates trajectories to match terrain geometry~\citep{comptonTerrainConsistentReferenceGuided2026}, motion-matching parkour chains retrieved clips before distilling a perceptive controller~\citep{wuPerceptiveHumanoidParkour2026}, and physics-consistent pipelines filter infeasible references with privileged rollouts~\citep{liOmniTrack2026}. We instead keep the raw clip fixed at deployment and use terrain-conformal references only as offline supervision for a raw-reference student. We also build on retargeting and trajectory optimization~\citep{choiNonparametricMotionRetargeting2020,villegasNeuralKinematicNetworks2018,zhangOmniRetarget2025,dantecWholeBodyTrajectoryOptimization2022,pajonWalkingGravel2017} and on teacher--student and residual-policy transfer~\citep{leeQuadrupedalRoughTerrain2020,kumarRMA2021,miki2022learning,heASAP2025,silverResidualPolicy2018,johannesResidualPolicy2019,zhaoResMimic2025,wangArmConstrained2024}.

\section{Method}
\label{sec:method}
Perceptive BFM is trained with the staged \emph{PMT} algorithm of Fig.~\ref{fig:teaser} (Alg.~\ref{alg:pmt-appendix}): TCRS synthesizes terrain-conformal references offline, a blind teacher is trained on them, a perceptive student that receives the raw reference and the elevation grid is distilled from the teacher, and the student is fine-tuned with PPO while the transferred tracking prior is updated conservatively. The raw reference remains the deployment command: terrain-conformal references supervise learning but are never supplied to the final policy.

\subsection{Problem Formulation}
\label{sec:problem}

Let $\mathbf{o}^{\mathrm{prop}}_t$ denote proprioception, $\mathbf{o}^{\mathrm{map}}_t$ the local elevation grid, and $\mathbf{m}^{\mathrm{raw}}_{t:t+H}$ a future window of the raw kinematic reference (target joint positions and velocities plus local motion-anchor displacements). Perceptive motion tracking asks for a policy
\begin{equation}
    \mathbf{a}_t =
    \pi_\theta\!\left(\mathbf{o}^{\mathrm{prop}}_t,
    \mathbf{o}^{\mathrm{map}}_t,
    \mathbf{m}^{\mathrm{raw}}_{t:t+H}\right),
    \label{eq:perceptive-tracking-policy}
\end{equation}
whose action preserves the behavior encoded in the reference while adapting contact, posture, and local feasibility to the terrain. The policy outputs a residual joint-position target
\begin{equation}
    \mathbf{q}^{\mathrm{pd}}_t = \mathbf{q}^{\mathrm{cmd}}_t + \mathbf{a}_t,
    \label{eq:residual-convention}
\end{equation}
where $\mathbf{q}^{\mathrm{cmd}}_t$ is the command-frame joint reference: $\mathbf{q}^{\mathrm{tcrs}}_t$ for the teacher and $\mathbf{q}^{\mathrm{raw}}_t$ for the deployed student.

\paragraph{Observations.}
The deployable actor receives projected gravity, base angular velocity, joint positions and velocities, previous actions, a 10-step proprioceptive history, and 21-step reference-command and motion-anchor displacement windows. Terrain perception is an \emph{elevation grid}, not an image: a torso-centered $17\times 11$ height grid over $1.6\,\mathrm{m}\times1.0\,\mathrm{m}$ at $0.1$\,m resolution with a validity mask, from a ray-cast scanner in simulation and from LiDAR elevation mapping on the robot (Sec.~\ref{sec:hardware}). Privileged terms are used only by the teacher, critic, or auxiliary losses; the deployed actor reads detached estimator outputs instead (App.~\ref{app:details}). Raw references are human clips retargeted to the robot by an in-house contact-grounded pipeline (App.~\ref{app:tcrs-details}).

\subsection{Terrain-Conformal Reference Synthesis}
\label{sec:adapter}

A raw human motion is behaviorally informative but not necessarily terrain-conformal. We therefore introduce a synthesis operator $\mathbf{m}^{\mathrm{tcrs}}_{1:T}=\mathcal{S}_{\mathrm{TCRS}}(\mathbf{m}^{\mathrm{raw}}_{1:T},\tau)$ that converts a raw clip and a terrain height field $h_\tau(x,y)$ into kinematic supervision for teacher training. TCRS does not solve contact-rich dynamics; it constructs contact-consistent, smooth, style-preserving references that make terrain adaptation learnable downstream (App.~\ref{app:tcrs-details}).

\paragraph{Contact-aware terrain reference.}
TCRS first estimates stance and swing intervals from foot height, velocity, and hysteresis thresholds; stance feet are latched to terrain support surfaces, while swing endpoints inherit the raw liftoff and landing timing. Horizontal placement is conservative: local support latching and edge penalties around the raw foot trace replace a global footstep planner, and toe/heel constraints are attenuated or rejected when support is unreliable. TCRS never re-times the clip: stance/swing intervals, footfall order, and global phase are inherited from the raw reference, so adaptation is spatially local and phase-preserving (its alignment requirement is measured in Sec.~\ref{sec:generality}).

\paragraph{Foot-geometry-aware swing optimization.}
For each foot $f$ and swing phase $s$, TCRS optimizes a virtual mid-foot point $\mathbf{p}^{\mathrm{mid}}_{f,t}$, the midpoint of the toe and heel offsets expressed in the world frame (Eq.~\eqref{eq:midfoot}), rather than the ankle origin, which balances toe and heel clearance near terrain discontinuities. The swing path is parameterized by $K$ \emph{control knots} $\mathbf{Y}=\{\mathbf{y}_k\}_{k=1}^{K}$, discrete waypoints that are spline-interpolated in time, with objective
\begin{equation}
\begin{aligned}
    J_s(\mathbf{Y};\tau)=&\; \lambda_{\mathrm{ref}}\sum_k\|\mathbf{y}_k-\mathbf{y}^{\mathrm{raw}}_k\|^2
    + \lambda_{\mathrm{sm}}\sum_k\|\Delta^2\mathbf{y}_k\|^2
    + \lambda_{\mathrm{clr}}\sum_k [h_\tau(x_k,y_k)+\delta-z_k]_+^2 \\
    &+ \lambda_{\mathrm{edge}}\Phi_{\mathrm{edge}}(\mathbf{Y};\tau)
    + \lambda_{\mathrm{end}}\|\mathbf{y}_{1}-\bar{\mathbf{y}}_{1}\|^2
    + \lambda_{\mathrm{end}}\|\mathbf{y}_{K}-\bar{\mathbf{y}}_{K}\|^2 ,
\end{aligned}
\label{eq:tcrs-swing-cost}
\end{equation}
where $[x]_+=\max(x,0)$. The terms preserve the raw swing, encourage smoothness, enforce a clearance margin $\delta$, discourage penetration near vertical faces ($\Phi_{\mathrm{edge}}$ penalizes toe/heel samples whose neighboring terrain queries indicate a vertical discontinuity within the foot footprint), and keep liftoff/landing endpoints fixed. The knots are optimized with batched sampling-based trajectory optimization (MPPI, Eq.~\eqref{eq:tcrs-sampling-update}), and the optimized mid-foot path is mapped back to ankle, toe, and heel targets for IK.

\paragraph{Support-aware root reconstruction.}
After foot replanning, TCRS reconstructs root height as the support-weighted average of the terrain height under each adapted foot plus the raw root-to-foot offset (Eq.~\eqref{eq:root-reconstruction}), clamped by leg reachability and smoothed across support transitions; during flight or weak-contact phases the filter falls back to the raw vertical profile with limited per-frame displacement.

\paragraph{Collision repair and multi-point contact IK.}
Finally, TCRS repairs lower-leg and foot collisions, attenuates unsupported toe/heel constraints near step edges, and solves a damped support-aware multi-point Jacobian IK problem over the twelve leg joints of the Unitree G1. TCRS retargets foot contacts only. Root translation is fixed to the reconstruction above; root orientation and upper-body joint references are preserved from the raw reference. The IK residual stacks ankle, toe, and heel point Jacobians with support-aware weights, posture and continuity regularization, penetration penalties, and damped least squares; multiseed fallback and continuity guards reject high-error or discontinuous branches (Eq.~\eqref{eq:ik-objective}). The output is the paired dataset $(\mathbf{q}^{\mathrm{raw}},\mathbf{q}^{\mathrm{tcrs}},\tau)$. We make no cross-embodiment claim: all instantiation and evaluation are on the G1.

\subsection{Policy Architecture and Training}
\label{sec:policy-architecture}

\paragraph{Blind teacher.}
The teacher is a Transformer actor--critic with tokenized proprioceptive history and reference-command windows (App.~\ref{app:architecture}); it receives TCRS references as commands and is trained with PPO to track adapted anchor, orientation, foot-position, and foot-velocity targets under energy and lateral foot/shin contact penalties.

\paragraph{Identity-gated perceptive student.}
The student inherits the same command/history Transformer and adds an elevation-grid encoder. Denote the pooled command-history latent by $\mathbf{u}_t$ and the terrain latent by $\mathbf{z}^{\mathrm{map}}_t=E_{\mathrm{map}}(\mathbf{o}^{\mathrm{map}}_t)$. Terrain affects the actor through two residual pathways with zero-initialized gates. The intent latent is modulated as
\begin{equation}
    \mathbf{u}'_t =
    \mathbf{u}_t + \tanh(\boldsymbol{\alpha}_u)
    \odot f_u(\mathbf{z}^{\mathrm{map}}_t);
    \label{eq:intent-gate}
\end{equation}
the actor trunk decodes the base action mean from the gated intent, $\boldsymbol{\mu}^{\mathrm{base}}_t=g_a(\mathbf{u}'_t,\mathbf{o}^{\mathrm{prop}}_t)$, and a second pathway adds a perception residual:
\begin{equation}
    \boldsymbol{\mu}_t =
    \boldsymbol{\mu}^{\mathrm{base}}_t +
    \tanh(\boldsymbol{\alpha}_a)
    \odot f_a([\mathbf{o}^{\mathrm{prop}}_t,\mathbf{z}^{\mathrm{map}}_t]).
    \label{eq:residual-action}
\end{equation}
At branch insertion, the gate vectors $\boldsymbol{\alpha}_u$ and $\boldsymbol{\alpha}_a$ are initialized to zero, while the residual MLPs retain standard initialization. The terrain contributions initially vanish, preserving the base tracker's outputs while allowing nonzero gate gradients.

\paragraph{Target-frame distillation and PPO fine-tuning.}
Because teacher and student act around different command frames, the student cannot imitate the teacher residual directly. We distill the teacher's effective PD target expressed relative to the raw reference,
\begin{equation}
    \mathbf{a}^{\star}_t =
    \left(\mathbf{q}^{\mathrm{tcrs}}_t + \boldsymbol{\mu}^{\mathrm{tea}}_t\right)
    - \mathbf{q}^{\mathrm{raw}}_t ,
    \label{eq:target-frame}
\end{equation}
and the student minimizes $\|\boldsymbol{\mu}^{\mathrm{stu}}_t-\mathbf{a}^{\star}_t\|_2^2$. During DAgger-style rollouts the teacher-control probability is annealed from $1$ to $0$; when the teacher controls the simulator, the applied action is the aligned target-frame action, in the residual units of Eq.~\eqref{eq:residual-convention} with the same scaling, limits, and clipping. PPO fine-tuning resumes from the distilled student, including its learned gates and residual branches, and uses raw-reference commands, elevation grids, and auxiliary estimator losses, with a lower learning-rate scale on the transferred backbone than on the terrain encoder, critic, and residual branches. Distillation thus initializes the terrain residuals, and fine-tuning refines them while contact and collision penalties discourage collapse back to flat-ground tracking. ``Resuming'' in our training logs refers to this post-distillation stage, not to extra optimization steps (App.~\ref{app:training-details}).

\section{Experiments}
\label{sec:experiments}
All quantitative results are \emph{simulation} results (IsaacLab; MuJoCo cross-check in Sec.~\ref{sec:deployed-eval}) for one fixed checkpoint, the publicly released deployment policy (project-page ONNX), whose weights also run on the robot; hardware evidence (Sec.~\ref{sec:hardware}) is presented as demonstrations of behavioral coverage.

\textbf{Setup} (App.~\ref{app:corpus-dr}, \ref{app:eval-protocol}). TCRS supervision is generated from the corpus of Sec.~\ref{sec:intro} on procedural terrain with stair risers $5$--$15$\,cm, height offsets to $0.25$\,m, support widths from $0.10$\,m, and slopes to $30^\circ$; clips are split before terrain pairing ($\approx100$ terrains per clip: terrain coverage, not new motion data), and $7.4\%$ of synthesized references ($\approx5\%$ of walk/run, $\approx20\%$ of flips) are filtered as teacher-untrackable. Rollouts evaluate five behaviors (walk, run, dance, backflip, cartwheel) on flat ground and on $11$ in-distribution cells: slopes S15--S60 (rise in cm over $2.5$\,m, $3.4$--$13.5^\circ$), boxes B5--B20 (cm), and six-step pyramid stairs P5--P15 (riser, cm), with the obstacle on the motion path, a random $\pm0.2$\,m start shift, and $30$ trials per cell, all attempts counted ($n=1{,}650$ terrain and $150$ flat trials per policy). Outcomes are mutually exclusive: \emph{completion}, \emph{fall}, and \emph{tracking loss} $=100-\text{completion}-\text{fall}$. Quality metrics use completed trials only: MPJPE against the raw reference (full body on flat ground; upper body on terrain, where the lower body must deviate and is scored against the teacher/TCRS target), lower-leg collision incidence, foot slip per contact, and PMT-normalized jerk. Baselines use three from-scratch seeds (spread $\le3.4$\,pp); bootstrap $95\%$ intervals on pooled completion are within $\pm4.5$\,pp, and a single $30$-trial cell carries about $\pm10$\,pp.

\subsection{Quality of Terrain-Conformal Reference Synthesis}
\label{sec:tcrs-quality}

\begin{wrapfigure}{r}{0.4\linewidth}
\centering
\vspace{-12pt}
\includegraphics[width=\linewidth]{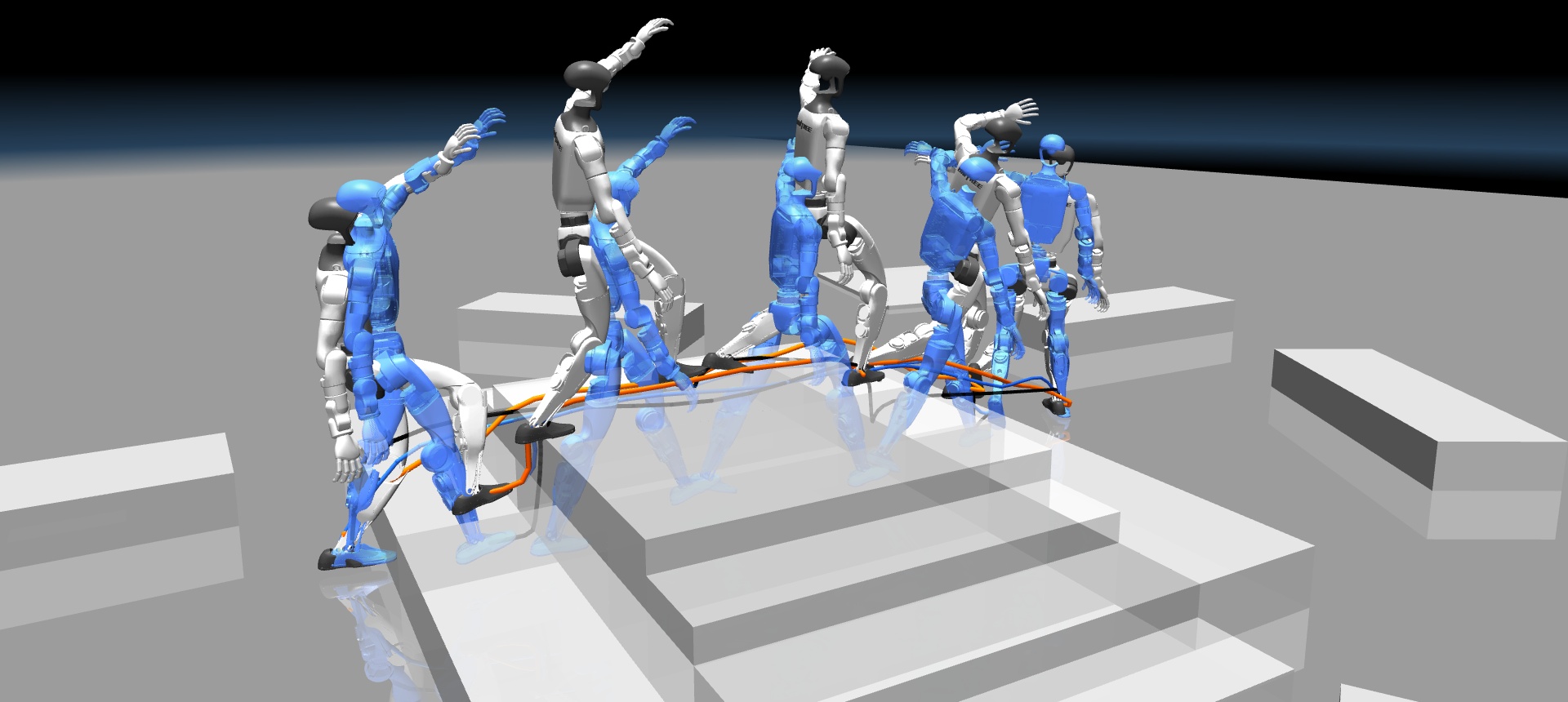}
\vspace{-14pt}
\caption{\small \textbf{TCRS trajectory synthesis.} Blue ghost: raw reference on terrain; opaque robot: TCRS output. Foot traces: MPPI mid-foot swing optimization of TCRS (orange), Cubic Interp (blue), direct $z$-lifting (black); the two saturations of each color are the left and right foot.}
\label{fig:tcrs-traj}
\vspace{-10pt}
\end{wrapfigure}
We evaluate TCRS against two simpler geometric edits, \emph{Z-offset} (lift the ankles to the local terrain height) and \emph{Cubic Interp+IK} (smooth a height-projected ankle path with cubic interpolation and solve single-point IK), with the reference-quality metrics of App.~\ref{app:tcrs-details} (Fig.~\ref{fig:tcrs-traj}). Over the full pool of $200$ clips with $20$ layouts each (mean $\pm$ std over clips), penetration depth is $1.7\pm1.2$\,cm for TCRS versus $4.1\pm2.6$\,cm for Z-offset and $1.9\pm1.3$\,cm for Cubic Interp+IK, and clearance violation is $5.7\%$ versus $24.9\%$ and $10.2\%$; the hardest slice, stepping stones with stairs over $30$ clips, shows the same pattern (Table~\ref{tab:tcrs-quality}). The robust gain of TCRS is \emph{clearance}, roughly halved relative to Cubic Interp+IK on both the slice and the pool; the penetration margin over Cubic Interp+IK is small and is not a headline. TCRS is not uniformly better: Cubic Interp+IK is smoother and marginally better in upper-body deviation, and TCRS has the highest float rate on the slice ($32.3\%$) and the pool ($26.1\%$), because near support transitions and step edges it prioritizes collision avoidance over exact contact. TCRS runs offline at about $2.3$\,s of compute per second of motion; Sec.~\ref{sec:ablation} tests whether these gains matter for control.

\subsection{Deployed-Checkpoint Evaluation Across Behaviors and Terrain Cells}
\label{sec:deployed-eval}

\begin{table}[t]
\centering
\small
\caption{\textbf{Deployed-checkpoint success rate (\%) per behavior and terrain cell, in simulation.} $30$ trials per cell, $\pm0.2$\,m start shifts, all attempts counted. S$x$: slope with $x$\,cm rise over $2.5$\,m; B$x$: $x$\,cm box; P$x$: $x$\,cm pyramid risers. Pooled: flat $100\%$, terrain $94.0\%$ ($1551/1650$). \rd{Red}: below $90\%$.}
\label{tab:grid}
\vspace{-4pt}
\setlength{\tabcolsep}{5pt}
\begin{tabular}{lcccccccccccc}
\toprule
 & Flat & S15 & S30 & S45 & S60 & B5 & B10 & B15 & B20 & P5 & P10 & P15 \\
\midrule
walk      & 100 & 100 & 100 & 100 & 100 & 97 & 97 & 100 & 100 & 100 & 100 & 100 \\
run       & 100 & 100 & 100 & 100 & \rd{83} & 97 & 97 & \rd{77} & \rd{60} & \rd{87} & \rd{73} & \rd{30} \\
dance     & 100 & 100 & 100 & 100 & 100 & 100 & 100 & 100 & 100 & 100 & 100 & 100 \\
backflip  & 100 & 100 & 100 & \rd{77} & \rd{83} & 100 & 100 & 100 & 100 & 100 & 100 & 100 \\
cartwheel & 100 & 100 & 100 & 93 & 100 & 100 & 100 & 100 & 100 & \rd{87} & 90 & \rd{43} \\
\bottomrule
\end{tabular}
\vspace{-2pt}
\end{table}

On every behavior--cell pair (Table~\ref{tab:grid}), the policy completes $100\%$ of flat-ground trials for all five behaviors and $94.0\%$ of the $1{,}650$ terrain trials, with a $2.4\%$ fall rate. Failures are structured: run degrades on tall boxes and risers (B20 $60\%$, P10 $73\%$, P15 $30\%$), cartwheel on the tallest risers (P15 $43\%$), and backflip on the steepest slopes (S45 $77\%$, S60 $83\%$), while walk and dance stay at $97$--$100\%$. The cells exclude sparse supports, recessed obstacles, and slopes above $13.5^\circ$ (training used up to $30^\circ$), so the campaign quantifies a subset of the trained terrain families (Sec.~\ref{sec:limitations}); transferring the same evaluation to MuJoCo lowers pooled terrain completion by $4.9$\,pp.

\subsection{Paired Baselines: TCRS Supervision and the Elevation Grid}
\label{sec:ablation}

\begin{table}[t]
\centering
\small
\caption{\textbf{Pipeline-matched baselines on identical tuples, in simulation} (three seeds each; terrain $=$ the $11$-cell pool). C.: completion (\%); Fall (\%) on terrain; MPJPE (mm): full-body on flat / upper-body on terrain; Coll.: lower-leg collision incidence (\%); Slip: cm per contact; Jerk: normalized to PMT. Quality columns use completed trials only ($n=322/386/360/1551$).}
\label{tab:paired}
\vspace{-4pt}
\setlength{\tabcolsep}{5pt}
\begin{tabular}{lccccccc}
\toprule
Policy & C.$_\text{flat}\!\uparrow$ & C.$_\text{terrain}\!\uparrow$ & Fall$\downarrow$ & MPJPE$_{\text{fb/ub}}\downarrow$ & Coll.$\downarrow$ & Slip$\downarrow$ & Jerk$\downarrow$ \\
\midrule
Blind   & 100 & 19.5 & 64.2 & 49\,/\,68 & 23.7 & 3.5 & 1.21 \\
No-TCRS & 100 & 23.4 & 58.7 & 46\,/\,72 & 34.2 & 2.6 & 1.33 \\
No-map  & 100 & 21.8 & 67.3 & 47\,/\,88 & 33.7 & 2.2 & 1.08 \\
\textbf{PMT (ours)} & \textbf{100} & \textbf{94.0} & \textbf{2.4} & \textbf{42\,/\,55} & \textbf{12.4} & \textbf{1.6} & \textbf{1.00} \\
\bottomrule
\end{tabular}
\vspace{-12pt}
\end{table}

Table~\ref{tab:paired} uses three baselines. \emph{Blind} trains the deployable architecture from scratch on raw references and reward, without the grid. \emph{No-TCRS} keeps the full teacher--student pipeline, architecture, and grid but replaces TCRS references and reward with raw ones, so only the supervision changes. \emph{No-map} uses TCRS references, teacher, and distillation but never sees the grid and is not fine-tuned.

\textbf{Correction to the submitted version.} The submitted paper reported that removing perception collapsed training reward to $3.6$, ``an order of magnitude below every other configuration.'' That comparison confounded the loss of perception with the loss of the distillation warm start (the full model resumed from it on the harder stepping-stone training terrain; the blind variant trained from scratch). Table~\ref{tab:paired} replaces it with pipeline-matched, \emph{evaluated} baselines; the earlier diagnostics are kept only in App.~\ref{app:logged-diagnostics} with that caveat.

\textbf{Findings.} The three baselines sit at a common floor of $19.5$--$23.4\%$ terrain completion (differences inside the $\pm4.5$\,pp interval, not to be ranked), and all complete flat ground, so the floor is a terrain-adaptation failure, not a tracking failure. The one clean single-factor contrast is No-TCRS\,$\to$\,PMT ($+70.6$\,pp): pipeline, architecture, and grid are identical and only references and reward change. The raw-motion reward penalizes exactly the deviations that terrain requires, so the grid receives little useful gradient; consistently, zeroing the grid at test time costs PMT $24.8$\,pp (Sec.~\ref{sec:probes}) but the No-TCRS policy about $0$\,pp. TCRS supervision without the grid (No-map) does not help either: the blind-distilled student fits a terrain-conditioned teacher it cannot observe and produces smooth (jerk $1.08$) but often invalid motion ($67.3\%$ falls). PMT also has the lowest collision incidence ($12.4\%$ vs.\ $23.7$--$34.2\%$), slip ($1.6$ vs.\ $2.2$--$3.5$\,cm), and upper-body MPJPE on terrain ($55$ vs.\ $68$--$88$\,mm): it preserves the commanded upper body while adapting the lower body.

\subsection{Does the Policy Use the Map? Test-Time Probes}
\label{sec:probes}

With the deployed checkpoint frozen and only its map input corrupted, pooled terrain completion falls from $94.0\%$ to $69.2\%$ with a zeroed map, $58.1\%$ with a spatially shuffled map, and $83.1\%$ with a map delayed by $200$\,ms. Shuffled is worse than zeroed: wrong terrain information hurts more than none, so the policy conditions on map content rather than using it as a regularizer. Corruptions in the training-noise regime cost at most $5$\,pp: per-cell Gaussian height noise with $\sigma\le0.04$\,m, up to $30\%$ cell dropout, or latency up to $100$\,ms. In a shifted-map probe, the foot-lift location moves with the map, and the terrain residual of Eq.~\eqref{eq:residual-action} departs from its flat-ground value about $0.34$\,s (${\approx}0.4$\,m at the clip's walking speed) before the commanded foot would reach the obstacle: the correction begins before contact.

\subsection{Teacher Tracking Against an Observation-Aligned SONIC Re-implementation}
\label{sec:teacher}

No public tracker consumes terrain-conformal references, so we re-implemented the SONIC architecture~\citep{luoSONIC2025} under our supervision, observations (with motion anchors and $10$-frame history), action space, and reward, and trained both models on the same clips and terrain with $24$ GPUs each (App.~\ref{app:sonic}, Table~\ref{tab:sonic}). Over $100$ clips per behavior and $300$-step rollouts, the teacher is better on $19$ of $25$ behavior--metric pairs and on all five means, e.g., anchor-position error $0.065$ vs.\ $0.212$\,m ($3.3\times$ lower; $2.1\times$ excluding run) and body-position error $0.046$ vs.\ $0.061$\,m; SONIC is better on body rotation and joint position for walk, dance, and cartwheel (pose/style cells). The teacher completes $96.5\%$ of the retained references. Terrain-aware parkour and navigation systems~\citep{zhuangDeepWholeBodyParkour2026,wuPerceptiveHumanoidParkour2026,comptonTerrainConsistentReferenceGuided2026} solve a different task (Sec.~\ref{sec:related-work}) and are not compared.

\subsection{Generality, Scaling, and Motion--Terrain Alignment}
\label{sec:generality}

\textbf{Source-held-out.} Retraining without the 100STYLE source and evaluating on the 100STYLE-clip tuples of the same cells gives $90.2\%$ completion, against $97.6\%$ for the full-data model on the same tuples (gap $7.4$\,pp). \textbf{Scaling.} Training on $1/4$, $1/2$, and $3/4$ of the corpus yields $78.9$, $87.0$, and $91.2\%$ pooled terrain completion (full: $94.0\%$), and $8/24/48$ GPUs yield $83.6/90.5/94.0\%$, monotone and not yet saturated. \textbf{Alignment envelope.} Because TCRS preserves global phase and PMT corrects timing only locally, phase-critical motions tolerate only a bounded offset from the terrain feature: sweeping offsets on a $0.1$ grid for the backflip and three further clips, success stays $\ge80\%$ within $\Delta x\in[-0.1,+0.1]$\,m and $\Delta t\in[-0.1,+0.1]$\,s, while locomotion clips are insensitive across the sweep. This is consistent with the $\pm0.2$\,m start shifts of Table~\ref{tab:grid}: those backflips include an approach that absorbs the shift, whereas the stair backflip of Fig.~\ref{fig:hero} is phase-critical and its hardware starts were aligned within the envelope.

\subsection{Real-Robot Deployment}
\label{sec:hardware}

\begin{wrapfigure}{r}{0.55\linewidth}
\centering
\vspace{-14pt}
\includegraphics[width=\linewidth]{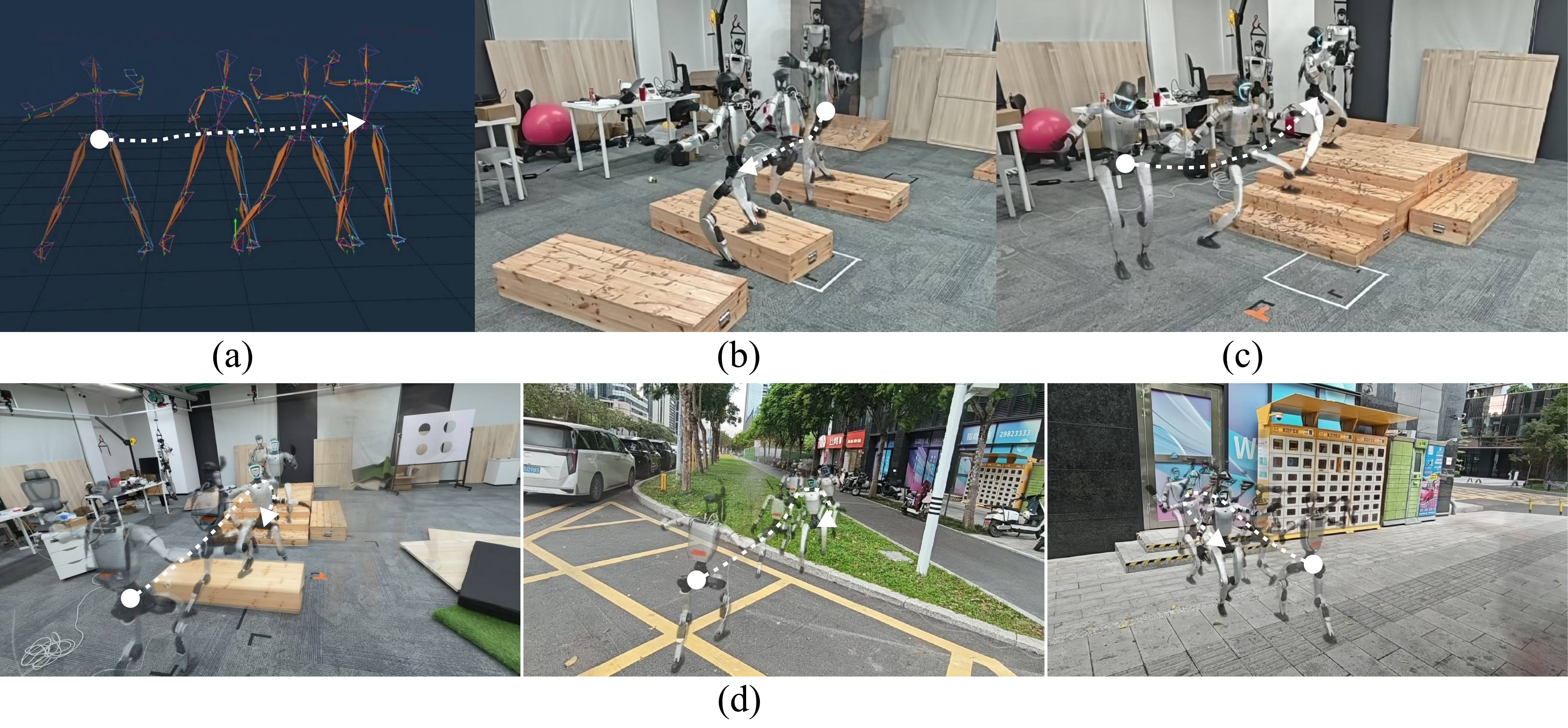}
\vspace{-14pt}
\caption{\small \textbf{Real-robot mocap mismatch.} (a) Human mocap on flat ground; (b,c) the robot tracks the (a) command over perceived obstacles; (d) walk-and-dance outdoors.}
\label{fig:mocap}
\vspace{-10pt}
\end{wrapfigure}
\textbf{Platform and perception.} The deployed policy runs on a $29$-DoF Unitree G1. The elevation grid is produced onboard from LiDAR measurements: LiDAR-inertial odometry and elevation mapping~\citep{mikiElevationMapping2022} accumulate a local map that is queried at $50$\,Hz into the same $17\times11$ grid and validity mask used in simulation; the actor consumes no RGB or depth image. The map is accumulated and odometry-dependent. Domain randomization (Table~\ref{tab:dr}) covers the sim-to-real gap; one checkpoint serves simulation and hardware with no per-motion tuning (output smoothing only for the live demo).

\textbf{Evidence.} We demonstrated $35$ motion classes on placed steps, boxes, and outdoor terrain. No repeated-trial hardware protocol was run, so these demonstrations establish behavioral coverage; all reported success rates are simulation results. Fig.~\ref{fig:hero} shows eight behavior--terrain pairs; Fig.~\ref{fig:mocap} isolates operator--environment mismatch: the human performs a mocap command on flat ground while the robot executes it over obstacles it perceives itself.

\section{Conclusion}
\label{sec:conclusion}
We presented Perceptive BFM, a motion-conditioned humanoid control framework that keeps the raw kinematic reference as the deployment command while a robot-centric elevation grid, entering through identity-gated residuals trained on TCRS supervision, corrects footholds, clearance, support height, and contact timing. On one fixed deployed checkpoint, the policy completes $94.0\%$ of $1{,}650$ in-distribution terrain trials in simulation where pipeline-matched baselines stay near $20\%$, probes show that it conditions on the terrain map, and the same checkpoint executes $35$ motion classes on hardware.

\section{Limitations}
\label{sec:limitations}
\begin{wrapfigure}{r}{0.33\linewidth}
\centering
\vspace{-16pt}
\includegraphics[width=\linewidth]{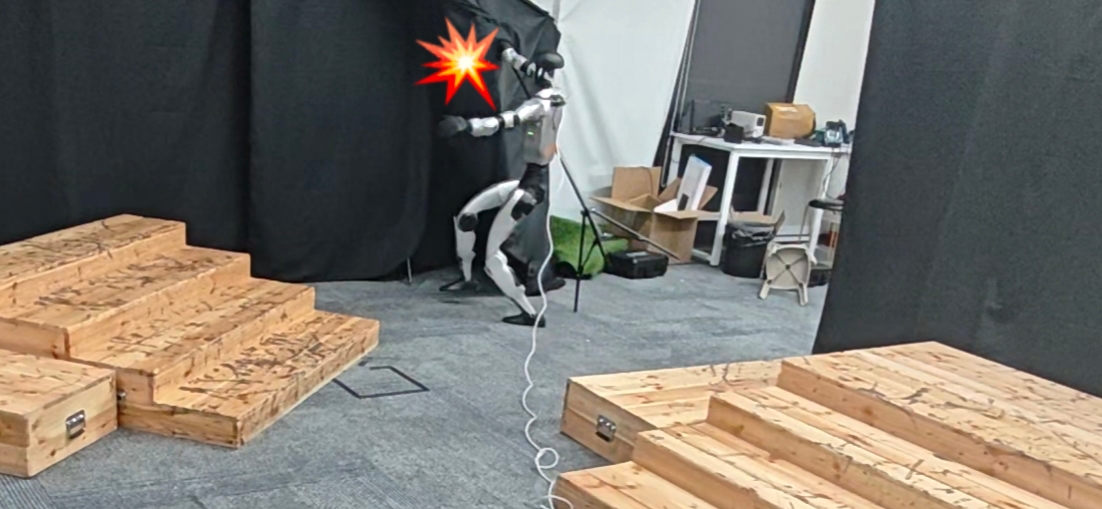}
\vspace{-14pt}
\caption{\small \textbf{Representative failure.} The upper-body command is collision-unaware, so arms or torso can strike obstacles.}
\label{fig:failure}
\vspace{-12pt}
\end{wrapfigure}
\textbf{Assumptions and scope of adaptation.} TCRS is a \emph{kinematic} synthesizer that assumes a static, rigid, observable height field; deformable, granular, slippery, or moving terrain is not modeled. TCRS retargets foot contacts and support height while preserving upper-body joint references. It does not retarget hand contacts or plan upper-body contacts, so arms or torso can strike nearby obstacles (Fig.~\ref{fig:failure}), and lower-leg collisions ($12.4\%$ incidence) are reduced but not eliminated.

\textbf{Alignment envelope and failure profile.} Phase-critical motions such as the stair backflip succeed only when the start is within about $\pm0.1$\,m and $\pm0.1$\,s of the terrain feature; such hardware demonstrations were started inside this envelope, and ``randomly placed'' terrain applies to locomotion and dance, not to phase-critical acrobatics. On in-distribution terrain, run completes only $60\%$ of trials on $20$\,cm boxes and $30\%$ on $15$\,cm risers, and cartwheel $43\%$ on $15$\,cm risers (Table~\ref{tab:grid}).

\textbf{Evaluation scope.} All success statistics are simulation results. The $11$-cell campaign excludes sparse supports, recessed obstacles, and slopes above $13.5^\circ$, so the trained terrain families are only partly quantified; sim-to-sim transfer to MuJoCo costs $4.9$\,pp. Hardware evidence consists of demonstrations of $35$ motion classes from one deployment campaign; repeated-trial hardware success rates are not reported.

\textbf{Motion quality, perception, and safety.} Some deployed videos show high-frequency jitter, small standing steps, and slow drift on uneven ground. Mitigating these motion-quality limitations remains future work. The accumulated elevation map depends on odometry. Occluded cells are handled only through the validity mask, latency near $200$\,ms degrades completion (Sec.~\ref{sec:probes}), and LiDAR-to-body calibration was not probed. Falls are physical risks; deployment uses in-policy fall recovery and an external watchdog (torque saturation, base-orientation excursions), and the policy must stay within the training terrain ranges, static obstacles, and the alignment envelope.

\clearpage

\bibliography{references}

\clearpage
\appendix
\section{Additional Implementation Details}
\label{app:details}
\subsection{PMT Training Algorithm}
\label{app:pmt-algorithm}

\begin{algorithm}[h]
\small
\caption{Perceptive Motion Tracking (PMT) training algorithm.}
\label{alg:pmt-appendix}
\begin{algorithmic}[1]
\State \textbf{Input:} retargeted raw motion clips $\mathcal{M}$, terrain generator $\mathcal{G}$, robot model, height scanner
\For{clip $m\in\mathcal{M}$ and terrain $\tau\sim\mathcal{G}$}
    \State synthesize $\mathbf{m}^{\mathrm{tcrs}}=\mathcal{S}_{\mathrm{TCRS}}(m,\tau)$ using Alg.~\ref{alg:tcrs-appendix}
    \State export paired raw/adapted references $(\mathbf{m}^{\mathrm{raw}},\mathbf{m}^{\mathrm{tcrs}},\tau)$
\EndFor
\State train blind teacher $\pi_T$ with PPO to track $\mathbf{m}^{\mathrm{tcrs}}$; discard references $\pi_T$ cannot track
\State compute target-frame labels from $\pi_T$ using Eq.~\eqref{eq:target-frame}
\State distill identity-gated perceptive student $\pi_S$ using raw-reference observations
\State fine-tune $\pi_S$ with PPO under raw-reference commands and elevation-grid observations
\State \textbf{Output:} deployable Perceptive BFM policy $\pi_S$
\end{algorithmic}
\end{algorithm}

\subsection{Training Stages}

Table~\ref{tab:task-map} maps each of the four PMT stages to its role in the pipeline.

\begin{table}[h]
\centering
\small
\caption{\textbf{PMT stage map.} Each stage in the PMT training algorithm corresponds to a distinct role in the pipeline.}
\label{tab:task-map}
\resizebox{\linewidth}{!}{%
\begin{tabular}{ll}
\toprule
Stage & Role \\
\midrule
TCRS data generation & Synthesizes paired raw/TCRS references on sampled terrain. \\
Blind teacher PPO & Tracks terrain-conformal references with privileged training groups. \\
Raw-reference distillation & Distills adapted-reference teacher targets into the raw-reference student frame. \\
Perceptive PPO fine-tuning & Fine-tunes the deployable student with raw commands and elevation grids. \\
\bottomrule
\end{tabular}%
}
\end{table}

\subsection{Observation Contract}

\begin{table}[h]
\centering
\small
\caption{\textbf{Observation groups used by PMT.} Privileged groups supervise critics or auxiliary heads but are not direct deployable actor inputs.}
\label{tab:obs-contract}
\resizebox{\linewidth}{!}{%
\begin{tabular}{llll}
\toprule
Group & Main contents & Shape or history & Deployment role \\
\midrule
Proprio & projected gravity, base angular velocity, joint pos/vel, previous action & 93D & actor input \\
Proprio history & unflattened proprioceptive history & 10 steps & actor temporal input \\
Command window & future reference velocity, gravity, joint command tokens & $21\times38$ & actor command input \\
Anchor delta window & local anchor displacement window & $21\times3$ & actor command input \\
Elevation grid & height cells plus validity mask & $17\times11$ cells, mask appended & terrain actor input \\
Critic & privileged reference/body/base information & current frame & training only \\
Auxiliary targets & base velocity, anchor, and foot-trajectory targets & current frame/window & auxiliary losses \\
\bottomrule
\end{tabular}%
}
\end{table}

\subsection{Reward and Optimization Details}

\begin{equation}
r_t = \sum_{k\in\{a,R,f,v\}} w_k\,e^{-\|\Delta_k\|^2/\sigma_k^2} - c_E E_t - c_C C_t .
\label{eq:reward-app}
\end{equation}
The PPO reward used by the adapted-reference teacher and the fine-tuning stage combines four exponential tracking terms with two penalties (Eq.~\eqref{eq:reward-app}). The tracking residuals are anchor position $\Delta_a=\mathbf{p}^A_t-\bar{\mathbf{p}}^A_t$, body orientation $\Delta_R=d_R(\mathbf{R}_t,\bar{\mathbf{R}}_t)$, and ankle position/velocity $\Delta_f,\Delta_v$ on $F$. $E_t$ is action/torque energy and $C_t$ is lateral foot/shin contact. Teacher rewards are evaluated against TCRS references, while student fine-tuning uses the raw-reference command frame with terrain-conditioned residual actions; the \emph{No-TCRS} baseline of Table~\ref{tab:paired} evaluates the same reward against the raw reference throughout the pipeline. Weights and standard deviations are listed in Table~\ref{tab:reward-details}; per-stage learning rates, entropy coefficients, and auxiliary-loss schedules in Table~\ref{tab:optimization-details}.

\begin{table}[h]
\centering
\small
\caption{\textbf{PPO reward terms for the teacher and fine-tuning tasks.} Exponential tracking terms use the listed standard deviation.}
\label{tab:reward-details}
\begin{tabular}{lll}
\toprule
Term & Weight & Notes \\
\midrule
Global anchor position tracking & $1.0$ & std $0.2$ m \\
Relative body orientation tracking & $0.5$ & std $0.35$ \\
Foot position tracking & $1.0$ & ankle bodies, std $0.1$ m \\
Foot linear-velocity tracking & $0.5$ & ankle bodies, std $1.0$ m/s \\
Energy & $-2{\times}10^{-5}$ & action/torque energy penalty \\
Foot/shin lateral contact & $-0.03$ & threshold 5 N on ankles and knees \\
\bottomrule
\end{tabular}
\end{table}

\begin{table}[h]
\centering
\small
\caption{\textbf{Training hyperparameters.} The fine-tuning stage starts from the distilled checkpoint and updates the inherited tracker more conservatively than the terrain branch.}
\label{tab:optimization-details}
\resizebox{\linewidth}{!}{%
\begin{tabular}{llll}
\toprule
Stage & Learning rate & Entropy & Additional losses or schedule \\
\midrule
Blind teacher PPO & $5{\times}10^{-4}$ & $0.005$ & 5 epochs, 4 minibatches, KL target $0.01$, velocity/anchor Huber losses \\
Distillation & $1{\times}10^{-4}$ & -- & MSE action loss, teacher-control mix annealed $1.0 \rightarrow 0.0$ \\
Perceptive PPO fine-tuning & $1{\times}10^{-4}$ & $0.001$ & backbone LR scale $0.3$, foot-trajectory Huber loss with delta $0.05$ \\
\bottomrule
\end{tabular}%
}
\end{table}

\subsection{Network and Data-Flow Summary}

\begin{figure}[t]
\centering
\includegraphics[width=0.98\linewidth]{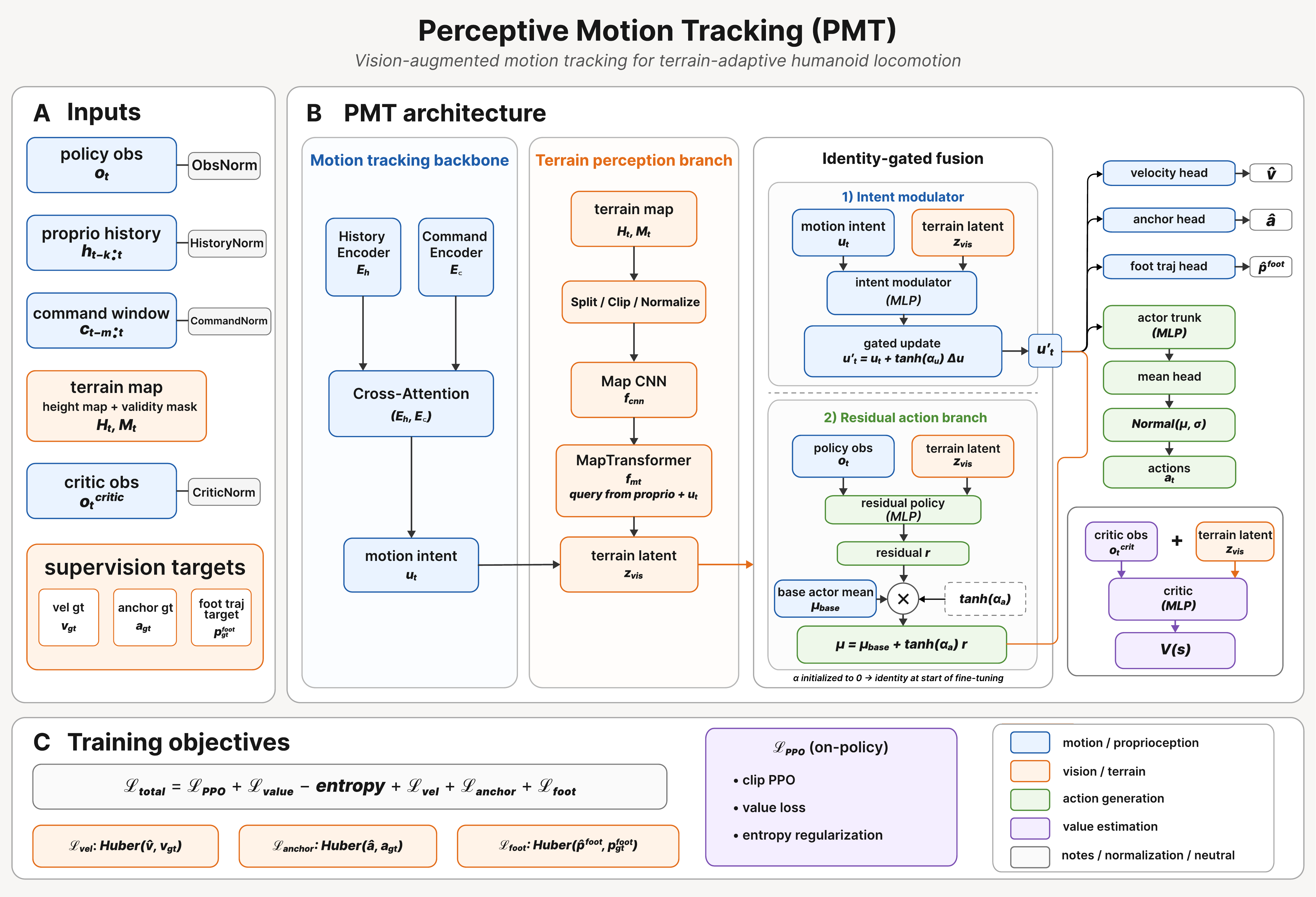}
\caption{\textbf{Detailed PMT network architecture.} (A) Inputs: policy observation $\mathbf{o}_t$, $10$-step proprioceptive history $\mathbf{h}_{t-k:t}$, command window $\mathbf{c}_{t-m:t}$, elevation grid and mask $(\mathbf{H}_t,\mathbf{M}_t)$, critic observation, and supervision targets. (B) PMT actor: a Transformer motion-tracking backbone with cross-attention encoders $E_h,E_c$ produces a motion intent $\mathbf{u}_t$; the terrain branch (map CNN $f_{\mathrm{cnn}}$ followed by a query-conditioned map Transformer $f_{\mathrm{mt}}$) produces a terrain latent ($z_{\mathrm{vis}}$ in the figure, $\mathbf{z}^{\mathrm{map}}_t$ in the text); the identity-gated fusion module updates the intent through $\mathbf{u}'_t=\mathbf{u}_t+\tanh(\alpha_u)\Delta\mathbf{u}$, the actor trunk decodes $\boldsymbol{\mu}^{\mathrm{base}}$ from $\mathbf{u}'_t$, and the residual branch adds $\tanh(\alpha_a)\,r$, with $\alpha$ initialized at zero and residual MLP weights retaining standard initialization, preserving the base tracker at branch insertion. Auxiliary heads predict base velocity, motion anchor, and foot trajectory; the critic operates on privileged inputs. (C) Training objective: PPO + value + entropy losses combined with Huber auxiliary losses on velocity, anchor, and foot trajectory. The ``vision'' wording in the figure denotes the elevation-grid branch; the actor consumes no image.}
\label{fig:architecture}
\end{figure}

\paragraph{Architecture.}
The actor consumes the raw reference command tokens and a 10-step proprioceptive history through the Transformer backbone, producing the motion-tracking latent. The $17\times11$ elevation grid together with its validity mask is processed by the terrain encoder (a map CNN followed by a query-conditioned map Transformer) into a terrain latent $\mathbf{z}^{\mathrm{map}}_t$ (written $z_{\mathrm{vis}}$ in Fig.~\ref{fig:architecture}). Two residual branches with zero-initialized gates inject this latent into the actor: an intent gate that modulates the pooled command-history latent (Eq.~\eqref{eq:intent-gate}) and an action-residual branch that adds to the action mean (Eq.~\eqref{eq:residual-action}). The gated intent $\mathbf{u}'_t$ is the input of the actor trunk that produces $\boldsymbol{\mu}^{\mathrm{base}}_t$; the residual branch then adds $\tanh(\boldsymbol{\alpha}_a)\odot f_a([\mathbf{o}^{\mathrm{prop}}_t,\mathbf{z}^{\mathrm{map}}_t])$. The actor output is a residual joint-position target around the raw reference, applied through Eq.~\eqref{eq:residual-convention}. The actor also reads two learned estimator heads driven by internal latents only (no privileged observation input): a base-velocity estimator and a motion-anchor-position estimator whose (detached) $3$D outputs are concatenated into the actor trunk, so the deployable actor never consumes ground-truth base velocity or anchor position. These estimators are supervised by the corresponding ground-truth targets as auxiliary losses only; a foot-trajectory head is likewise supervised against the teacher target and is not an actor input. The future motion-anchor displacement window remains a direct command-token input, not an estimated quantity.

\paragraph{Identity initialization.}
When the gated terrain branches are introduced, $\boldsymbol{\alpha}_u=\mathbf{0}$ and $\boldsymbol{\alpha}_a=\mathbf{0}$, while the intent and action residual MLPs retain standard weight initialization. Thus their gated contributions are initially zero, but their outputs can provide nonzero gradients to the gates. In the ungated residual ablation, the final linear layers of these branches are instead zero-initialized to preserve the same initial base outputs. PPO fine-tuning loads the distilled student checkpoint, including its learned gate values and residual-branch weights; it does not reinitialize them.

\paragraph{Data flow.}
Each retargeted raw clip is paired with sampled terrains to form $(m^{\mathrm{raw}},\tau)$ tuples, which TCRS converts into terrain-conformal references $m^{\mathrm{tcrs}}$ (Alg.~\ref{alg:tcrs-appendix}). The blind teacher is trained with PPO on $m^{\mathrm{tcrs}}$. During distillation, the teacher's action is converted into the raw-reference frame using Eq.~\eqref{eq:target-frame}, and the raw-reference student is fit by MSE. The converted action uses the same scaled residual joint-position convention as the student action and is clipped by the same residual limits before it is used for imitation or teacher-control rollouts. The student is then fine-tuned with PPO under raw-reference commands and elevation-grid observations. At deployment, the policy receives only the raw reference, proprioception, and the onboard elevation grid; TCRS is never queried online.

\subsection{Terrain-Conformal Reference Synthesis Details}
\label{app:tcrs-details}

\paragraph{Retargeting.}
Raw references are human motion clips retargeted to the G1 by an in-house contact-grounded retargeting pipeline: a differentiable kinematics model (PyTorch) is fit to the human joint trajectories under joint limits, and foot contacts detected in the source motion are grounded through MuJoCo forward simulation of the retargeted pose. TCRS consumes the retargeted clip; it does not modify the retargeting.

\paragraph{Mid-foot frame, sampling update, and root reconstruction.}
For each foot $f$, the virtual mid-foot point used by the swing optimizer is the midpoint of the toe and heel offsets $\mathbf{r}^{\mathrm{toe}},\mathbf{r}^{\mathrm{heel}}$ in the ankle frame,
\begin{equation}
    \mathbf{r}^{\mathrm{mid}}=\tfrac{1}{2}(\mathbf{r}^{\mathrm{toe}}+\mathbf{r}^{\mathrm{heel}}),\qquad
    \mathbf{p}^{\mathrm{mid}}_{f,t}=\mathbf{p}^{\mathrm{ankle}}_{f,t}+\mathbf{R}_{f,t}\mathbf{r}^{\mathrm{mid}} .
    \label{eq:midfoot}
\end{equation}
The control knots of Eq.~\eqref{eq:tcrs-swing-cost} are optimized with batched sampling-based trajectory optimization; with temporally smoothed perturbations $\epsilon^{(j)}$ and temperature $\eta$, each iteration applies the softmin-weighted update
\begin{equation}
    \mathbf{Y}\leftarrow \mathbf{Y} + \sum_j
    \frac{\exp(-J_s(\mathbf{Y}+\epsilon^{(j)})/\eta)}
    {\sum_\ell \exp(-J_s(\mathbf{Y}+\epsilon^{(\ell)})/\eta)}\epsilon^{(j)} .
    \label{eq:tcrs-sampling-update}
\end{equation}
After foot replanning, the target root height is the support-weighted average over feet of the terrain height under the adapted foot plus the raw root-to-foot offset: for support weights $w_{f,t}$ and adapted foot positions $(x^{\mathrm{tcrs}}_{f,t},y^{\mathrm{tcrs}}_{f,t})$,
\begin{equation}
    z^{\star}_{\mathrm{root},t}
    = \frac{\sum_f w_{f,t}
    \left[h_\tau(x^{\mathrm{tcrs}}_{f,t},y^{\mathrm{tcrs}}_{f,t})
    + z^{\mathrm{raw}}_{\mathrm{root},t}-z^{\mathrm{raw}}_{f,t}\right]}
    {\sum_f w_{f,t}+\epsilon} ,
    \label{eq:root-reconstruction}
\end{equation}
clamped by leg reachability and smoothed across support transitions; during flight or weak-contact phases the filter falls back to the raw vertical profile with limited per-frame displacement.

\paragraph{Multi-point contact IK objective.}
Each frame, TCRS solves a damped support-aware multi-point Jacobian IK problem over the twelve leg joints, using foot-contact targets only. Root translation is fixed to the support-aware reconstruction stage, while root orientation and upper-body joint references are preserved from the raw reference. For the foot point set $\mathcal{P}=\{\mathrm{ankle},\mathrm{toe},\mathrm{heel}\}$, the local IK update solves
\begin{equation}
\begin{aligned}
    \Delta\mathbf{q}^{\mathrm{leg},\star}
    = \arg\min_{\Delta\mathbf{q}^{\mathrm{leg}}}\;
    &\sum_{f\in\{L,R\}}\sum_{p\in\mathcal{P}}
    \left\|\mathbf{W}_{f,p}\left(\mathbf{J}_{f,p}\Delta\mathbf{q}^{\mathrm{leg}}-\mathbf{e}_{f,p}\right)\right\|^2
    +\lambda_{\mathrm{post}}\|\mathbf{q}^{\mathrm{leg}}-\mathbf{q}^{\mathrm{raw,leg}}\|^2 \\
    &+\lambda_{\mathrm{cont}}\|\mathbf{q}^{\mathrm{leg}}-\mathbf{q}^{\mathrm{prev,leg}}\|^2
    +\lambda_{\mathrm{pen}}\Psi_{\mathrm{pen}}(\mathbf{q}^{\mathrm{leg}})
    +\lambda_{\mathrm{dls}}\|\Delta\mathbf{q}^{\mathrm{leg}}\|^2,
\end{aligned}
\label{eq:ik-objective}
\end{equation}
where $\Delta\mathbf{q}^{\mathrm{leg}}$ contains the leg-joint increments, $\mathbf{e}_{f,p}$ are ankle/toe/heel residuals, $\mathbf{W}_{f,p}$ are support-aware point weights, and $\Psi_{\mathrm{pen}}$ penalizes foot or lower-leg penetration. Multiseed fallback and continuity guards reject high-error or discontinuous branches.

\begin{algorithm}[h]
\small
\caption{Terrain-Conformal Reference Synthesis (TCRS).}
\label{alg:tcrs-appendix}
\begin{algorithmic}[1]
\State \textbf{Input:} raw clip $m^{\mathrm{raw}}$, terrain height field $h_\tau$, contact offsets, swing-optimization parameters, IK weights
\State detect contact masks using foot height, speed, and hysteresis thresholds
\State build stance references by anchoring support feet to terrain surfaces
\For{each foot $f$ and swing phase $s=[t_0,t_1]$}
    \State convert ankle targets to a mid-foot frame using Eq.~\eqref{eq:midfoot}
    \State initialize trajectory knots from the raw mid-foot path
    \For{sampling iteration $i=1,\ldots,N_{\mathrm{iter}}$}
        \State sample temporally smoothed knot perturbations
        \State evaluate tracking, smoothness, clearance, vertical-face, and endpoint costs using Eq.~\eqref{eq:tcrs-swing-cost}
        \State update knots with the softmin-weighted sampling update in Eq.~\eqref{eq:tcrs-sampling-update}
    \EndFor
    \State convert optimized mid-foot trajectory back to ankle, toe, and heel targets
\EndFor
\State reconstruct root height from support contacts using Eq.~\eqref{eq:root-reconstruction}
\State repair lower-leg and foot collisions; attenuate unsupported toe/heel point weights near edges
\For{frame $t=1,\ldots,T$}
    \State solve multi-point Jacobian IK using Eq.~\eqref{eq:ik-objective}
    \State retry with secondary seeds if foot error, penetration, or joint jumps exceed thresholds
\EndFor
\State \textbf{Output:} terrain-conformal reference $m^{\mathrm{tcrs}}$, contact masks, realized feet/root, and diagnostics
\end{algorithmic}
\end{algorithm}

\paragraph{Reference-quality metrics.}
We evaluate TCRS before policy rollout, following the principle that a reference generator should be tested independently from the tracker it trains. Let
\begin{equation}
    d_\tau(\mathbf{p}) = p_z - h_\tau(p_x,p_y)
\end{equation}
be signed terrain clearance. Terrain penetration is
\begin{equation}
    E_{\mathrm{pen}} =
    \frac{1}{T|\mathcal{B}|}\sum_{t=1}^{T}\sum_{b\in\mathcal{B}}
    [ - d_\tau(\mathbf{p}_{b,t})]_+,
\end{equation}
where $\mathcal{B}$ contains foot and lower-leg points. Stance contact error and floating rate are
\begin{equation}
\begin{aligned}
    E_{\mathrm{contact}} &=
    \frac{1}{|\mathcal{C}|}\sum_{(f,t)\in\mathcal{C}}
    |d_\tau(\mathbf{p}_{f,t})|, \qquad
    R_{\mathrm{float}} =
    \frac{1}{|\mathcal{C}|}\sum_{(f,t)\in\mathcal{C}}
    \mathbb{I}[d_\tau(\mathbf{p}_{f,t})>\epsilon_{\mathrm{float}}].
\end{aligned}
\end{equation}
Swing-clearance violation is
\begin{equation}
    R_{\mathrm{clear}} =
    \frac{1}{|\mathcal{S}|}\sum_{(f,t)\in\mathcal{S}}
    \mathbb{I}[d_\tau(\mathbf{p}_{f,t})<c_{\min}],
\end{equation}
and style deviation is measured by upper-body deviation from the raw clip,
\begin{equation}
    E_{\mathrm{upper}} =
    \frac{1}{T|\mathcal{J}_{ub}|}
    \sum_{t,j\in\mathcal{J}_{ub}}
    \|\mathbf{x}^{\mathrm{tcrs}}_{j,t}-\mathbf{x}^{\mathrm{raw}}_{j,t}\|_2 .
\end{equation}
Foot smoothness is measured as mean finite-difference foot acceleration,
\begin{equation}
    E_{\mathrm{smooth}} =
    \frac{1}{(T-2)|\mathcal{F}|}
    \sum_{t=2}^{T-1}\sum_{f\in\mathcal{F}}
    \left\|
    \frac{\mathbf{p}_{f,t+1}-2\mathbf{p}_{f,t}+\mathbf{p}_{f,t-1}}{\Delta t^2}
    \right\|_2 .
\end{equation}
A good synthesizer should reduce penetration and clearance violations, avoid excessive nominal-contact floating, keep $E_{\mathrm{upper}}$ small, and maintain smooth foot trajectories. Table~\ref{tab:tcrs-quality} reports the $30$-clip stepping-stones-with-stairs slice; the pooled statistics in Sec.~\ref{sec:tcrs-quality} use $200$ clips with $20$ sampled layouts each and report mean $\pm$ standard deviation over clips.

\begin{table}[h]
\centering
\small
\caption{\textbf{TCRS reference quality on the stepping-stones-with-stairs slice} ($30$ clips, before any rollout). Lower is better; bold marks the best per column. Pooled $200\times20$ statistics are in the text.}
\label{tab:tcrs-quality}
\setlength{\tabcolsep}{4.5pt}
\resizebox{\linewidth}{!}{%
\begin{tabular}{lccccc}
\toprule
Reference & Pen.\ depth (cm) $\downarrow$ & Float rate (\%) $\downarrow$ & Clear.\ viol.\ (\%) $\downarrow$ & Foot smooth.\ (m/s$^2$) $\downarrow$ & Upper dev.\ (cm) $\downarrow$ \\
\midrule
Z-offset (FK projection) & $5.48$          & $\mathbf{12.4}$ & $33.8$         & $15.1$         & $6.51$ \\
Cubic Interp + IK        & $2.69$          & $31.7$          & $14.3$         & $\mathbf{6.9}$ & $\mathbf{3.98}$ \\
TCRS (ours)              & $\mathbf{2.38}$ & $32.3$          & $\mathbf{7.4}$ & $8.6$          & $4.00$ \\
\bottomrule
\end{tabular}%
}
\end{table}

\paragraph{Reference filtering.}
After teacher training, references on which the teacher fails to track are removed before distillation: $7.4\%$ of the synthesized references overall, about $5\%$ of walk/run clips and about $20\%$ of flip clips. The teacher completes $96.5\%$ of the retained references.

\paragraph{Implementation specifics.}
TCRS preserves the root orientation and upper-body joint references from the raw clip. Root translation is shaped by the support-aware reconstruction stage (Eq.~\eqref{eq:root-reconstruction}) and then held fixed during IK, so the IK solver updates only the twelve leg joints using ankle, toe, and heel targets. The multi-point structure enables foot-geometry-aware contact near step edges. TCRS preserves upper-body joint posture during synthesis, although root translation changes its world-coordinate trajectory; the deployed student tracks the same upper-body command. Hand-contact retargeting is not part of TCRS.

\subsection{Network Architecture}
\label{app:architecture}

\subsection{Training Corpus and Domain Randomization}
\label{app:corpus-dr}

\paragraph{Motion corpus.}
The pretraining corpus contains $16{,}813$ retargeted clips ($\approx43.3$\,h) from LAFAN1~\citep{harveyLAFAN2020}, AMASS~\citep{mahmoodAMASS2019}, 100STYLE~\citep{mason100STYLE2022}, SNAP, and our own motion capture, covering walking, running, turning, side-stepping, expressive walk--dance sequences, and acrobatic motions including hand-contact skills. Clips are split at the clip level before terrain pairing; each clip is then paired with about $100$ sampled terrains, which adds terrain coverage rather than unique motion data (gains saturate near $100$ pairings per clip). Each motion is exported at $30$\,fps and paired with sampled height fields as described in Sec.~\ref{sec:adapter}. The source-held-out experiment of Sec.~\ref{sec:generality} retrains the full pipeline with every 100STYLE clip removed and evaluates on the 100STYLE-clip tuples of the same terrain cells.

\paragraph{Control and simulation.}
Training uses IsaacLab/IsaacSim with a physics step of $\Delta t=5$\,ms and control decimation $4$, giving a $50$\,Hz policy rate; the configured episode horizon is $15$\,s, or $750$ policy steps. Each variant runs $6144$ environments per GPU across $48$ A800 GPUs.

\paragraph{Domain randomization.}
To support sim-to-real transfer we randomize, per episode, the parameters in Table~\ref{tab:dr}. Terrain perception is corrupted by additive per-cell height noise and a planar drift to emulate elevation-map error, and the robot is periodically pushed by direct base-velocity perturbation.

\begin{table}[h]
\centering
\small
\caption{\textbf{Domain randomization ranges used during training.} Values are sampled per episode unless noted.}
\label{tab:dr}
\begin{tabular}{lll}
\toprule
Parameter & Range & Notes \\
\midrule
Static friction & $[0.3,\,1.6]$ & per-body material \\
Dynamic friction & $[0.3,\,1.2]$ & per-body material \\
Base CoM offset & $\pm0.025$ m & per axis \\
Default joint position & $\pm0.1$ rad & reset perturbation, prob.\ $0.2$ \\
Push velocity & $\pm0.2$ m/s & periodic base push \\
Elevation-grid noise & $\pm0.06$ m & additive, per cell, clipped \\
\bottomrule
\end{tabular}
\end{table}

\subsection{Deployed-Checkpoint Evaluation Protocol}
\label{app:eval-protocol}

\paragraph{Checkpoint identity.}
All policy results in Sec.~\ref{sec:experiments} use one checkpoint, the deployment policy exported to ONNX and released on the project page; the same weights run on the robot. Baseline policies are trained with three independent seeds each; the table reports the seed mean, and the largest seed spread on pooled terrain completion is $3.4$\,pp.

\paragraph{Tuples and episodes.}
A tuple is a (behavior, cell, trial) triple. Each cell is a single obstacle placed on the motion path of the clip: slopes S15/S30/S45/S60 rise $15$/$30$/$45$/$60$\,cm over a $2.5$\,m run ($3.4$/$6.8$/$10.2$/$13.5^\circ$), boxes B5/B10/B15/B20 are $5$--$20$\,cm high, and pyramid stairs P5/P10/P15 have six steps with $5$/$10$/$15$\,cm risers. For each trial the start pose is shifted uniformly within $\pm0.2$\,m along and across the path. An episode runs the full clip horizon. Completion means reaching the full clip horizon without triggering a non-timeout termination term. A rollout is unsuccessful when any configured termination term for reference-relative anchor position, anchor orientation, or tracked end-effector position is triggered; non-foot contact alone is not the failure criterion. Unsuccessful trials are reported as \emph{fall} or \emph{tracking loss}; these three outcome categories are mutually exclusive and sum to $100\%$. All attempts are counted; nothing is discarded or re-run.

\paragraph{Metrics.}
MPJPE is the mean per-joint position error in mm between the robot and the raw reference after aligning the anchor frame. On flat ground it is computed over all joints. On terrain it is computed over the upper-body joints only, because the lower body must deviate from the raw reference to realize terrain contacts; lower-body realization is instead scored against the teacher/TCRS target and by foothold realization (fraction of TCRS stance contacts realized on the intended support). Collision incidence is the fraction of episodes with at least one lower-leg/terrain contact above the $5$\,N contact threshold; it is not an episode outcome and is reported separately from the outcome budget. Slip is the mean planar displacement of a stance foot per contact in cm. Jerk is the mean norm of the third finite difference of joint positions, reported relative to the PMT value on the same tuples. Quality metrics are averaged over completed trials only, so for baselines that complete mainly the easiest cells they describe an easier subset ($n=322$, $386$, $360$ completed terrain trials for Blind, No-TCRS, No-map versus $1{,}551$ for PMT).

\paragraph{Uncertainty.}
Pooled completion intervals are clip-resampled paired bootstrap $95\%$ intervals over the $1{,}650$ terrain trials and are within $\pm4.5$\,pp; the mutual differences among the three baselines in Table~\ref{tab:paired} fall inside this interval. A single $30$-trial cell has a binomial $95\%$ half-width of about $10$\,pp near $80\%$ success, so per-cell values in Table~\ref{tab:grid} should be read qualitatively.

\paragraph{Sim-to-sim cross-check.}
Re-running the same tuples in MuJoCo with the same checkpoint lowers pooled terrain completion by $4.9$\,pp relative to IsaacLab.

\paragraph{Map-corruption probes.}
Probes in Sec.~\ref{sec:probes} replace only the elevation-grid input of the frozen checkpoint on the same tuples: \emph{zeroed} sets every cell to the flat-ground height; \emph{shuffled} permutes cells spatially within the grid; \emph{delayed} feeds the grid from $200$\,ms earlier; the noise, dropout, and latency sweeps use per-cell Gaussian height noise, random invalidation of cells through the validity mask, and delays up to $100$\,ms. The shifted-map probe translates the grid along the path and records when the action residual of Eq.~\eqref{eq:residual-action} first departs from its flat-ground value relative to the commanded contact time.

\paragraph{Alignment envelope.}
For the backflip and three further clips, the terrain feature is offset spatially by $\Delta x$ and the command temporally by $\Delta t$ on a $0.1$ grid ($\Delta x$ in m, $\Delta t$ in s and as a phase fraction of the clip), with $30$ trials per grid point; the envelope is the region with $\ge80\%$ completion.

\subsection{SONIC Re-implementation}
\label{app:sonic}

\begin{table}[h]
\centering
\small
\caption{\textbf{Teacher tracking error, ours\,/\,SONIC re-implementation.} Same TCRS clips, terrain, supervision, observations (SONIC receives the motion anchors and the $10$-frame history), and action space; $24$ GPUs each; $100$ clips per behavior, $300$-step rollouts. $E_{\mathrm{bp}}/E_{\mathrm{ap}}$: body/anchor position (m); $E_{\mathrm{br}}/E_{\mathrm{ar}}$: body/anchor rotation (rad); $E_{\mathrm{jp}}$: joint position (rad). Bold $=$ better; \rd{red} $=$ ours worse ($6/25$).}
\label{tab:sonic}
\setlength{\tabcolsep}{6pt}
\begin{tabular}{lccccc}
\toprule
Behavior & $E_{\mathrm{bp}}$ & $E_{\mathrm{ap}}$ & $E_{\mathrm{br}}$ & $E_{\mathrm{ar}}$ & $E_{\mathrm{jp}}$ \\
\midrule
walk      & \textbf{.033}/.037 & \textbf{.040}/.078 & \rd{.099}/\textbf{.086} & \textbf{.039}/.043 & \rd{.493}/\textbf{.419} \\
run       & \textbf{.083}/.135 & \textbf{.143}/.682 & \textbf{.349}/.551 & \textbf{.163}/.255 & \textbf{1.428}/2.087 \\
dance     & \textbf{.038}/.042 & \textbf{.045}/.079 & \rd{.151}/\textbf{.116} & \textbf{.053}/.055 & \rd{.728}/\textbf{.582} \\
backflip  & \textbf{.027}/.037 & \textbf{.033}/.080 & \textbf{.121}/.127 & \textbf{.058}/.073 & \textbf{.540}/.626 \\
cartwheel & \textbf{.048}/.056 & \textbf{.065}/.139 & \rd{.291}/\textbf{.218} & \textbf{.104}/.126 & \rd{1.305}/\textbf{1.020} \\
\midrule
mean      & \textbf{.046}/.061 & \textbf{.065}/.212 & \textbf{.202}/.219 & \textbf{.083}/.110 & \textbf{.899}/.947 \\
\bottomrule
\end{tabular}
\end{table}

Table~\ref{tab:sonic} gives the per-behavior errors summarized in Sec.~\ref{sec:teacher}. The comparison re-implements the SONIC tracker architecture~\citep{luoSONIC2025} inside our training stack so that both models receive the same supervision (TCRS references on the same terrain), the same observation contract (including the motion-anchor inputs and the $10$-frame proprioceptive history, i.e., observation-aligned), the same action space, and the same reward. Each model is trained with $24$ GPUs; the reported checkpoints are our teacher at iteration $9{,}800$ and the SONIC re-implementation at iteration $9{,}999$. Tracking errors are averaged over $100$ clips per behavior and $300$-step rollouts. Body-position and anchor-position errors are in meters, body-rotation and anchor-rotation errors in radians, and joint-position error in radians. The run behavior dominates the anchor-position gap ($0.143$ vs.\ $0.682$\,m); excluding run the anchor ratio is $2.1\times$. The SONIC re-implementation is stronger on body-rotation and joint-position error for walk, dance, and cartwheel.

\subsection{Training-Time Component Diagnostics}
\label{app:logged-diagnostics}

\begin{table}[h]
\centering
\small
\caption{\textbf{End-of-training logged rollout diagnostics on stepping stones with stairs.} All variants share the task, reward, observation contract, action space, and $10$k-iteration final-PPO budget on $48$ A800 GPUs; values are mined from the final $100$ logged training iterations, and success is $\mathrm{total\_success}/(\mathrm{total\_success}+\mathrm{total\_failed})$ from the training logger. These are training-time diagnostics on a harder terrain family than Sec.~\ref{sec:deployed-eval}, not held-out evaluations; stage ablations that remove the distillation warm start (\emph{w/o TCRS}, \emph{w/o map}, \emph{w/o target-frame align}) train from scratch by construction, so reward is not comparable across rows and is left unbolded.}
\label{tab:ablation}
\begin{tabular}{lcccc}
\toprule
Variant & Succ.\ (\%) $\uparrow$ & Reward $\uparrow$ & Anchor err.\ (m) $\downarrow$ & Joint err.\ (rad) $\downarrow$ \\
\midrule
\textbf{PMT (full)}        & $\mathbf{55.1}$ & $45.2$ & $\mathbf{0.159}$ & $\mathbf{1.243}$ \\
w/o identity gate          & $30.4$ & $14.9$ & $0.231$ & $1.428$ \\
map-concat (no gate)       & $29.9$ & $14.5$ & $0.234$ & $1.421$ \\
w/o TCRS (raw ref.)        & $27.3$ & $17.9$ & $0.229$ & $1.833$ \\
w/o target-frame align     & $26.7$ & $9.3$  & $0.221$ & $1.959$ \\
w/o map (blind)            & $26.5$ & $10.6$ & $0.238$ & $2.464$ \\
\bottomrule
\end{tabular}
\end{table}

\begin{figure}[h]
\centering
\includegraphics[width=0.6\linewidth]{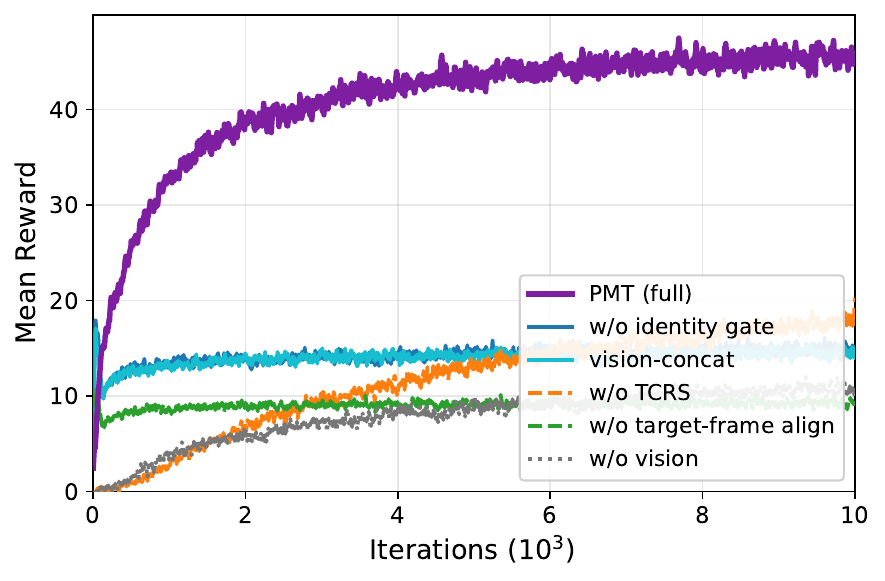}
\caption{\small \textbf{Training reward diagnostics} for the six variants of Table~\ref{tab:ablation}. The full PMT fine-tuning stage resumes from the distilled checkpoint while the stage ablations train from scratch, so absolute reward is not comparable across variants; the curve is a coarse optimization diagnostic only.}
\label{fig:training-curves}
\end{figure}

Table~\ref{tab:ablation} and Fig.~\ref{fig:training-curves} retain the training-time diagnostics of the submitted version for the component choices that were not re-run under the deployed-checkpoint protocol: the identity gate versus a plain residual (\emph{w/o identity gate}) or feature concatenation (\emph{map-concat}), and target-frame alignment versus distilling the teacher's native residual (\emph{w/o target-frame align}). They are logged on the harder stepping-stones-with-stairs training terrain, where a blind policy terminates quickly, which is why the \emph{w/o map} row here and the \emph{Blind} row of Table~\ref{tab:paired} differ in absolute level. The submitted version's ``training reward collapses to $3.6$'' claim for the blind variant was drawn from these logs and is superseded by Sec.~\ref{sec:ablation}: it conflated removing perception with removing the distillation warm start. Within this table, removing the identity gate or replacing it with concatenation lowers logged success from $55.1\%$ to about $30\%$, and removing target-frame alignment lowers logged reward to $9.3$ with elevated joint error, consistent with the teacher target being harder to optimize when it is not expressed in the raw-reference action frame; without seed-level variance, the $0.2$--$0.8$\,pp differences among the lowest rows should not be ranked.

\subsection{Real-Robot Deployment Protocol}
\label{app:hardware-protocol}

\paragraph{Platform and perception.}
The deployable policy runs on a 29-DoF Unitree G1 humanoid. Robot-centric terrain perception uses onboard LiDAR measurements. LiDAR-inertial odometry estimates the body pose, and an elevation-mapping module~\citep{mikiElevationMapping2022} accumulates the LiDAR returns into a local, body-centered elevation map. At each control step ($50$\,Hz), the map is queried on the $17\times11$ grid ($1.6\,\mathrm{m}\times1.0\,\mathrm{m}$, $0.1$\,m cells) used in training, and cells without valid measurements are flagged through the validity mask. The grid is expressed in the same torso-centered frame as in simulation, so the observation contract is preserved across the sim-to-real transfer. The actor consumes no RGB or depth image. The map is accumulated over time and depends on odometry; occluded regions behind steps are filled only once observed, and unobserved cells are masked. LiDAR-to-body calibration error was not probed.

\paragraph{Indoor protocol.}
Indoor trials place rectangular boxes, raised steps, artificial-turf surfaces, and mixed obstacle layouts in a motion-capture-equipped lab. Each terrain configuration is exercised with multiple commanded behaviors from the broad behavior set, including walking, running, side-walking, backward stepping, dance, and acrobatic motions. Terrain is placed randomly for locomotion and dance commands; for phase-critical acrobatics (the stair backflip), the start is aligned with the terrain feature within the envelope of Sec.~\ref{sec:generality}.

\paragraph{Outdoor protocol.}
Outdoor trials cover stairs, grass, isolated steps, recessed flower beds, and sidewalk transitions. Hardware safety is supervised by a runtime watchdog that triggers a soft fall-over recovery if the estimator detects torque saturation or a base-orientation excursion outside the training-time envelope; the policy additionally has in-policy fall recovery.

\paragraph{Mocap mismatch protocol.}
The mocap mismatch setup deliberately separates the human and robot environments: the human performs the commanded motion on flat ground while the robot executes the corresponding kinematic command over placed steps and cube obstacles. This protocol directly tests whether robot-centric terrain perception can ground a human command that contains no matching terrain information; the live commands were never seen in training.

\paragraph{Nature of the evidence.}
The same deployed checkpoint is used across all behaviors and terrains, with no per-command reward tuning, no per-terrain controller switching, and only output smoothing added for the live demonstration. Over the deployment campaign, $35$ motion classes were executed on placed terrain. Because several behaviors are high-energy and contact-rich (e.g., backflips on raised blocks), we did not run a repeated-trial campaign per behavior--terrain pair. The hardware results are therefore demonstrations of behavioral coverage (Figs.~\ref{fig:hero} and~\ref{fig:mocap}); all success rates in the paper are simulation results.

\subsection{Training and Compute Details}
\label{app:training-details}

All policy variants share the same task definition, reward terms (Eq.~\eqref{eq:reward-app}), observation contract (Table~\ref{tab:obs-contract}), action space, episode horizon, environment count, and PPO hyperparameters. Every stage is trained on $48$ NVIDIA A800 GPUs; the final PPO stage of each variant runs $10$k iterations, and the full PMT pipeline additionally uses TCRS data generation, adapted-reference teacher training, and raw-reference distillation before fine-tuning. ``Resuming'' in the training logs of the full model refers to this post-distillation fine-tuning stage, which starts from the distilled checkpoint; the total number of PPO iterations is matched across the variants of Table~\ref{tab:paired}, each of which is trained with three seeds. The scaling runs of Sec.~\ref{sec:generality} keep this schedule and vary either the fraction of the corpus ($1/4$, $1/2$, $3/4$) or the number of GPUs ($8$, $24$, $48$) with the environment count per GPU fixed.

\paragraph{Motion-quality note.}
Some deployed videos show high-frequency jitter, small standing steps, and slow drift on uneven ground. Mitigating these motion-quality limitations remains future work.

\end{document}